\documentclass[10pt,letterpaper]{article}

\usepackage{xspace}
\usepackage{times}
\usepackage{epsfig}
\usepackage{graphicx}
\usepackage{fullpage}
\usepackage{fancyhdr}
\usepackage{amsmath}
\usepackage{amssymb}
\usepackage{subfigure}
\usepackage{algorithm}
\usepackage{algorithmic}
\usepackage{threeparttable}
\newcommand{\argmin}{\operatornamewithlimits{argmin}}
\newcommand{\argmax}{\operatornamewithlimits{argmax}}

\makeatletter
\DeclareRobustCommand\onedot{\futurelet\@let@token\@onedot}
\def\@onedot{\ifx\@let@token.\else.\null\fi\xspace}

\def\etal{\emph{et al}\onedot}

\fancypagestyle{plain}{
\fancyhf{}

\rhead{Technical Report TR-2011-11 \\
University of British Columbia}}
\begin{document}

\title{Local Naive Bayes Nearest Neighbor for Image Classification}

\author{Sancho McCann \qquad David G. Lowe \\
{Department of Computer Science, University of British Columbia} \\
{2366 Main Mall, Vancouver, BC, Canada V6T 1Z4}}

\maketitle

\begin{abstract}
  We present Local Naive Bayes Nearest Neighbor, an improvement to the
  NBNN image classification algorithm that increases classification
  accuracy and improves its ability to scale to large numbers of
  object classes. The key observation is that only the classes
  represented in the local neighborhood of a descriptor contribute
  significantly and reliably to their posterior probability
  estimates. Instead of maintaining a separate search structure for
  each class, we merge all of the reference data together into one
  search structure, allowing quick identification of a descriptor's
  local neighborhood. We show an increase in classification accuracy
  when we ignore adjustments to the more distant classes and show that
  the run time grows with the log of the number of classes rather than
  linearly in the number of classes as did the original. This gives a
  100 times speed-up over the original method on the Caltech 256
  dataset. We also provide the first head-to-head comparison of NBNN
  against spatial pyramid methods using a common set of input
  features. We show that local NBNN outperforms all previous NBNN
  based methods and the original spatial pyramid model. However, we
  find that local NBNN, while competitive with, does not beat
  state-of-the-art spatial pyramid methods that use local soft
  assignment and max-pooling.
\end{abstract}

\section{Introduction}
\begin{figure}
\centering
\subfigure[The original NBNN asks, ``Does this descriptor look like a keyboard? a car? ... a dog?'']{
  \includegraphics[width=0.5\linewidth]{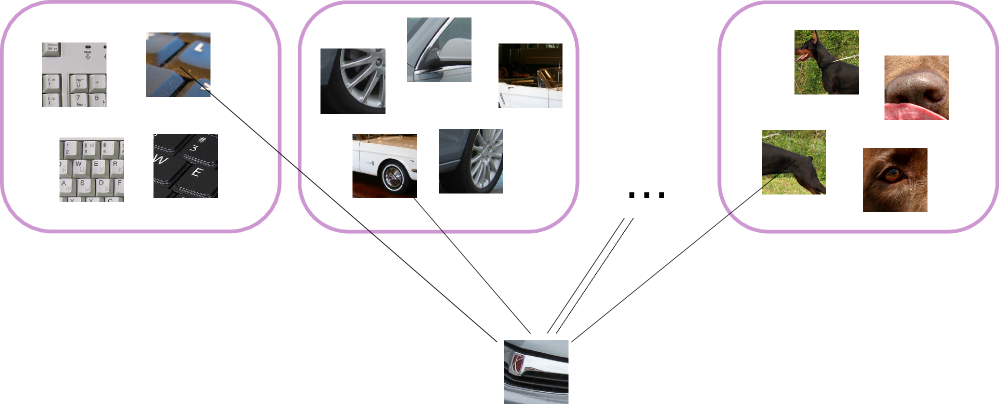}
}
\subfigure[Local NBNN asks, ``What does this descriptor look like?'']{
  \includegraphics[width=0.5\linewidth]{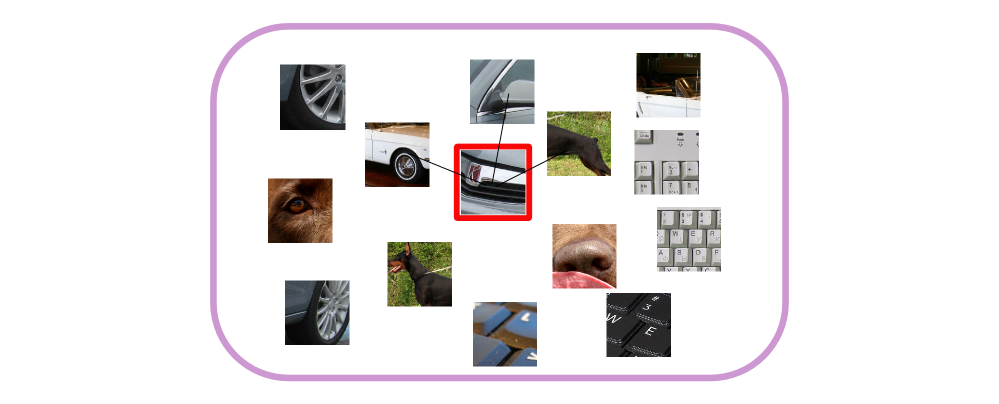}
}
\caption{Instead of considering classes individually, we search one
  merged index.}
\label{fig:method}
\end{figure}
A widely used approach to object category recognition has been the
bag-of-words method \cite{Csurka2004} combined with the spatial
pyramid match kernel \cite{Lazebnik2006}. This approach uses visual
feature extraction, quantizes features into a limited set of visual
words, and performs classification, often with a support vector
machine \cite{Jurie2005, Lampert2009}.

In contrast to the bag-of-words method, Boiman \etal \cite{Boiman2008}
introduced a feature-wise nearest neighbor algorithm called
\textit{Naive Bayes Nearest Neighbor} (NBNN). They do not quantize the
visual descriptors and instead retain all of the reference descriptors
in their original form.

Boiman \etal\cite{Boiman2008} showed that quantizing descriptors in
the bag-of-words model greatly decreases the discriminativity of the
data. The bag-of-words model usually reduces the high dimensional
feature space to just a few thousand visual words.

Despite NBNN's independence assumption (independence of the
descriptors in the query image), Boiman \etal demonstrated
state-of-the-art performance on several object recognition datasets,
improving upon the commonly used SVM classifier with a spatial pyramid
match kernel.

NBNN is a simple algorithm. The task is to determine the most probable
class $\hat{C}$ of a query image $Q$. Let $d_1, \ldots, d_n$ be all
the descriptors in the query image. The training data for a class is a
collection of descriptors extracted from a set of labelled example
images. These are stored in data structures that allow for efficient
nearest neighbor searches (the nearest neighbor of descriptor $d_i$ in
class $C$ is $\operatorname{NN}_C(d_i)$). The original NBNN is listed
as Algorithm \ref{alg:NBNN}.

\newcommand{\nbnn}{\ensuremath{\mbox{\sc NBNN}}}
\begin{algorithm}[h]
  \caption{$\nbnn(Q)$ \cite{Boiman2008}}\label{alg:nbnn}
\begin{algorithmic}
\REQUIRE A nearest neighbor index for each $C$, queried using
${\rm NN}_C()$.
\REQUIRE A query image $Q$, with descriptors $d_i$.
\medskip
\FORALL{descriptors $d_i \in Q$}
  \FORALL{classes $C$}
    \STATE ${\rm totals}[C] \gets {\rm totals}[C] + \|d_i - {\rm NN}_C(d_i)\|^2$
  \ENDFOR
\ENDFOR
\RETURN $\argmin_C {\rm totals}[C]$
\end{algorithmic}
\label{alg:NBNN}
\end{algorithm}

Our contribution is a modification to the original NBNN algorithm that
increases classification accuracy and provides a significant speed-up
when scaling to large numbers of classes. We eliminate the need to
search for a nearest neighbor in each of the classes. Instead, we
merge the reference datasets together and use an alternative nearest
neighbor search strategy in which we only adjust the scores for the
classes nearest to any query descriptor. The question becomes, ``What
does this descriptor look like?'', instead of ``Does this descriptor
look like one from a car? a duck? a face? a plane? ...'' Figure
\ref{fig:method} gives a conceptual visualization.

We also provide the first head-to-head comparison of NBNN based
methods with spatial pyramid methods using a common feature
set. Previous work \cite{Boiman2008, Tuytelaars2011} has only provided
comparisons with published figures while extracting different feature
sets for their experiments.

\section{Relation to previous work}
An obvious issue with the naive Bayes approach is that it makes
the unrealistic assumption that image features provide independent
evidence for an object category.

In defense of the naive Bayes assumption, Domingos and Pazzani
\cite{Domingos1997} demonstrate the applicability of the naive Bayes
classifier even in domains where the independence assumption is
violated. They show that while the independence assumption
\textit{does} need to hold in order for the naive Bayes classifier to
give optimal probability estimates, the classifier can perform well as
regards misclassification rate even when the assumption doesn't
hold. They perform extensive evaluations on many real-world datasets
that violate the independence assumption and show classification
performance on par with or better than other learning methods.

Behmo \etal\cite{Behmo2010} corrects NBNN for the case of unbalanced
training sets. Behmo \etal implemented and compared a variant of NBNN
that used $n$ 1-vs-all binary classifiers, highlighting the effect of
unbalanced training data. In the experiments we present, the training
sets are approximately balanced, and we compare our results to the
original NBNN algorithm. Behmo \etal also point out that a major
practical limitation of NBNN is the time that is needed to perform the
nearest neighbor search, which is what our work addresses.

The most recent work on NBNN is by Tuytelaars \etal
\cite{Tuytelaars2011}. They use the NBNN response vector of a query
image as the input features for a kernel SVM. This allows for
discriminative training and combination with other complimentary
features by using multiple kernels. The kernel NBNN gives increased
classification accuracy over using the basic NBNN algorithm. Our work
is complimentary to this in that the responses resulting from our
local NBNN could also be fed into their second layer of discriminative
learning. Due to the poor scaling of the original NBNN algorithm,
Tuytelaars \etal had to heavily subsample the query images in order to
obtain timely results for their experiments, hampering their absolute
performance values. In NBNN, what dominates is the time needed to
search for nearest neighbors in each of the object category search
structures. Even approximate methods can be slow here and scale
linearly with the number of categories.

The method we will introduce is a local nearest neighbor modification
to the original NBNN. Other methods taking advantage of local coding
include locality constrained linear coding by \cite{Wang2010} and
early cut-off soft assignment by \cite{Liu2011}. Both limit themselves
to using only the local neighborhood of a descriptor during the coding
step. By restricting the coding to use only the local dictionary
elements, these methods achieve improvements over their non-local
equivalents. The authors hypothesize this is due to the manifold
structure of descriptor space, which causes Euclidean distances to
give poor estimates of membership in codewords far from descriptor
being coded \cite{Liu2011}.

NBNN methods can be compared to the popular spatial pyramid methods
\cite{Boureau2010b, Lazebnik2006, Wang2010}, which achieve
state-of-the-art results on image categorization problems. The
original spatial pyramid method used hard codeword assignment and
average pooling within each of the hierarchical histogram bins. Today,
the best performing variants of spatial pyramid use local coding
methods combined with max pooling \cite{Boureau2010b, Boureau2011,
  Liu2011, Wang2010}. State-of-the-art spatial pyramid methods achieve
high accuracy on benchmark datasets, but there has been no
head-to-head comparison of NBNN methods against spatial pyramid
methods. Previous work has only compared against published figures,
but these comparisons are based on different feature sets, which makes
it difficult to isolate the contributions of the features from the
contributions of the methods.

\section{Naive Bayes Nearest Neighbor}
\label{sec:nbnn}
To help motivate and justify our modifications to the NBNN algorithm,
this section provides an overview of the original derivation
\cite{Boiman2008}. Each image $Q$ is classified as belonging to class
$\hat{C}$ according to
\begin{equation}
\hat{C} = \argmax_C p(C|Q).
\end{equation}

Assuming a uniform prior over classes and applying Bayes' rule,
\begin{equation}
\hat{C} = \argmax_C \log(p(Q|C)).
\end{equation}
The assumption of independence of the descriptors $d_i$ found in image
$Q$ gives
\begin{align}
\hat{C} & = \argmax_C\left[\log(\prod_{i=1}^np(d_i|C))\right] \\
& = \argmax_C\left[\sum_{i=1}^n\log p(d_i|C)\right]. \label{eqn:sum-of-log-probs}
\end{align}
Next, approximating $p(d_i|C)$ in Equation \ref{eqn:sum-of-log-probs}
by a Parzen window estimator, with kernel $K$, gives
\begin{equation}
\hat{p}(d_i|C) = \frac{1}{L}\sum_{j=1}^L K(d_i - d_j^C)
\end{equation}
where there are $L$ descriptors in the training set for class $C$ and
$d_j^C$ is the $j$-th nearest descriptor in class $C$. This can be
further approximated by using only the $r$ nearest neighbors as
\begin{equation}
\hat{p}_r(d_i|C) = \frac{1}{L}\sum_{j=1}^r K(d_i - d_j^C)
\end{equation}
and NBNN takes this to the extreme by using only the single
nearest neighbor ($\operatorname{NN}_C(d_i)$):
\begin{equation}
\hat{p}_1(d_i|C) = \frac{1}{L} K(d_i - \operatorname{NN}_C(d_i)). \label{eqn:single-nearest-neighbor}
\end{equation}
Choosing a Gaussian kernel for $K$ and substituting Equation
\ref{eqn:single-nearest-neighbor} (the single nearest neighbor
approximation of $p(d_i|C)$) into Equation \ref{eqn:sum-of-log-probs}
(the sum of log probabilities) gives:
\begin{align}
\hat{C} & = \argmax_C\left[\sum_{i=1}^n\log \frac{1}{L} e^{-\frac{1}{2\sigma^2}\|d_i - \operatorname{NN}_C(d_i)\|^2} \right] \\
& = \argmin_C\left[\sum_{i=1}^n\|d_i - \operatorname{NN}_C(d_i)\|^2 \right]. \label{eqn:nbnn}
\end{align}
Equation \ref{eqn:nbnn} is the NBNN classification rule: find the
class with the minimum distance from the query image.

\section{Towards local NBNN}
\label{sec:towards}
Before introducing local NBNN, we first present some results
demonstrating that we can be selective with the updates that we choose
to apply for each query descriptor. We start by re-casting the NBNN
updates as adjustments to the posterior log-odds of each class. In
this section, we show that only the updates giving positive evidence
for a class are necessary.

The effect of each descriptor in a query image $Q$ can be expressed as
a log-odds update. This formulation is useful because it allows us to
restrict updates to only those classes for which the descriptor gives
significant evidence. Let $C$ be some class and $\overline{C}$ be the
set of all other classes.

The odds ($\mathcal{O}$) for class $C$ is given by
\begin{align}
\mathcal{O}_C & = \frac{P(C|Q)}{P(\overline{C}|Q)} \\
& = \frac{P(Q|C)P(C)}{P(Q|\overline{C})P(\overline{C})} \\
& = \prod_{i=1}^n\frac{P(d_i|C)}{P(d_i|\overline{C})}\frac{P(C)}{P(\overline{C})}.
\end{align}

Taking the log and applying Bayes' rule again gives:
\begin{align}
\log(\mathcal{O}_C) & = \sum_{i=1}^n\log\frac{P(d_i|C)}{P(d_i|\overline{C})} + \log\frac{P(C)}{P(\overline{C})} \\
& = \sum_{i=1}^n\underbrace{\log\frac{P(C|d_i)P(\overline{C})}{P(\overline{C}|d_i)P(C)}}_\text{increment} + \underbrace{\log\frac{P(C)}{P(\overline{C})}}_{\text{prior}}. \label{eqn:log-odds}
\end{align}

Equation \ref{eqn:log-odds} has an intuitive interpretation. The prior
odds are $P(C)/P(\overline{C})$. Each descriptor then contributes a
change to the odds of a given class determined by the posterior odds
of $C$ given $d_i$, $P(C|d_i)/P(\overline{C}|d_i)$, and how they
differ from the prior odds as seen in the increment term of Equation
\ref{eqn:log-odds}. If the posterior odds are equal to the prior odds,
$\text{increment} = 0$, if the posterior odds are greater than the
prior odds, $\text{increment} > 0$, and if the posterior odds are less
than the prior odds, the $\text{increment} < 0$.

This allows an alternative classification rule that's expressed in
terms of log-odds increments:
\begin{equation}
\hat{C} = \argmax_C \left[\sum_{i=1}^n \log\frac{P(C|d_i)P(\overline{C})}{P(\overline{C}|d_i)P(C)} + \log\frac{P(C)}{P(\overline{C})} \right] \label{eqn:log-increments}
\end{equation}
where the prior term can be dropped if you assume equal class
priors. The increment term is simple to compute if we leave $P(C|d_i)
\propto e^{-\|d_i - \operatorname{NN}_C(d_i)\|^2}$ as in the original.

The benefit that comes from this formulation is that we can be
selective about which increments to actually use: we can use only the
significant log-odds updates. For example, we can decide to only
adjust the class posteriors for which the descriptor gives a positive
contribution to the sum in Equation \ref{eqn:log-increments}. Table
\ref{tab:thresholding} shows that this selectivity does not affect
classification accuracy.

\begin{table}\footnotesize
\begin{center}
\begin{tabular}{|c|c|c|}
\hline
Method & Avg. \# increments & Accuracy \%\\
\hline\hline
Full NBNN & 101 & 55.2 $\pm0.97$\\
Positive increments only & 55.0 & 55.6 $\pm1.17$\\
\hline
\end{tabular}
\end{center}
\caption{Effect of restricting increments to only the positive
  increments on a downsampled version (128x128) of the Caltech 101
  dataset. The $\pm$ shows one standard deviation.}
\label{tab:thresholding}
\end{table}

\section{Local NBNN}
\label{sec:merging}
The selectivity introduced in the previous section shows that we do
not need to update each class's posterior for each descriptor. This
section shows that by focusing on a much smaller, local neighborhood
(rather than on a particular log-odds threshold), we can use an
alternate search strategy to speed up the algorithm, and also achieve
better classification performance by ignoring the distances to classes
far from the query descriptor.

Instead of performing a search for a query descriptor's nearest
neighbor in each of the classes' reference sets, we search for only
the nearest few neighbors in a single, merged dataset comprising all
the features from all labelled training data from all classes. Doing
one approximate k-nearest-neighbor search in this large index is much
faster than querying each of the classes' approximate-nearest-neighbor
search structures. This is a result of the sublinear growth in
computation time with respect to index size for approximate nearest
neighbor search algorithms as discussed in Section \ref{sec:ann}. This
allows the algorithm to scale up to handle many more classes, avoiding
a prohibitive increase in runtime.

\begin{figure}
\centering
\includegraphics[width=0.5\linewidth]{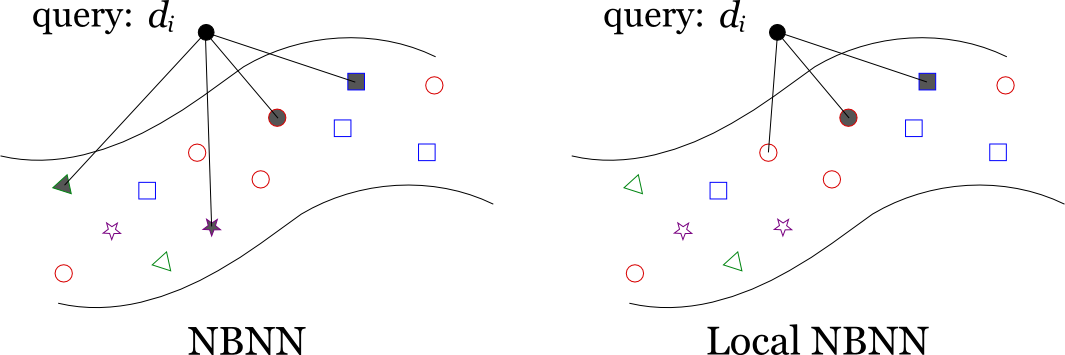}
\caption{NBNN finds the nearest neighbor from each of the classes
  (the shapes, in this figure). Local NBNN
  retrieves only the local neighborhood, finding nearest neighbors
  from only some of the classes. The shaded descriptors are those that
  would be used for updating the distance totals. We only use the
  closest member from any class, and don't find an example from each
  class.}
\label{fig:LNBNN}
\end{figure}

This is an approximation to the original method. For each test
descriptor in a query image, we do not find a nearest neighbor from
every class, only a nearest neighbor from classes that are represented
in the $k$ nearest descriptors to that test descriptor. We call this
local NBNN, visualized in Figure \ref{fig:LNBNN}.

It is important to properly deal with the set of background classes
which were not found in the $k$ nearest neighbors of $d$. To handle
the classes that were \textit{not} found in the $k$ nearest neighbors,
we conservatively estimate their distance to be the distance to the
$k+1$-st nearest neighbor (this can be thought of as an upper bound on
the density of background features). In practice, instead of adjusting
the distance totals to every class, it is more efficient to only
adjust the distances for the relatively few classes that were found in
the $k$ nearest neighbors, but discount those adjustments by the
distance to background classes (the $k$+1st nearest neighbor). This
does not affect the minimum.

The local NBNN algorithm is as follows:

\newcommand{\lnbnn}{\ensuremath{\mbox{\sc LocalNBNN}}}
\begin{algorithm}[h]
  \caption{$\lnbnn(Q, k)$}\label{alg:lnbnn}
\begin{algorithmic}
\REQUIRE A nearest neighbor index comprising all descriptors, queried using ${\rm NN}({\it descriptor}, {\it \# neighbors})$.
\REQUIRE A class lookup, ${\rm Class}({\it descriptor})$ that returns the class of a descriptor.
\medskip
\FORALL{descriptors $d_i \in Q$}
  \STATE $\{p_1, p_2, \ldots, p_{k+1}\} \gets {\rm NN}(d_i, k+1)$
  \STATE ${\it dist}_B \gets \|d_i - p_{k+1}\|^2$
  \FORALL{categories $C$ found in the $k$ nearest neighbors}
    \STATE ${\it dist}_C = \min_{\{p_j | {\rm Class}(p_j) = C\}} \|d_i - p_j\|^2$
    \STATE ${\rm totals}[C] \gets {\rm totals}[C] + {\it dist}_C - {\it dist}_B$
  \ENDFOR
\ENDFOR
\RETURN $\argmin_C {\rm totals}[C]$
\end{algorithmic}
\end{algorithm}

\subsection{Approximate nearest neighbors and complexity}
\label{sec:ann}
Our algorithm scales with the log of the number of classes rather than
linearly in the number of classes. This analysis depends on the nearest
neighbor search structure that we use.

For both our implementation of the original NBNN and local NBNN, we
use FLANN \cite{Muja2009} to store descriptors in efficient
approximate nearest neighbor search structures. FLANN is a library for
finding approximate nearest neighbor matches that is able to
automatically select from among several algorithms and tune their
parameters for a specified accuracy. It makes use of multiple,
randomized KD-trees as described by Silpa-Anan and Hartley
\cite{Silpa-Anan2008} and is faster than single KD-tree methods like
ANN \cite{Arya1998} (used by Boiman \etal in the original NBNN) or
locality sensitive hashing methods. The computation required and the
accuracy of the nearest neighbor search is controlled by the number of
leaf nodes checked in the KD-trees.

Following the analysis by Boiman \etal \cite{Boiman2008}, let $N_T$ be
the number of training images per class, $N_C$ the number of classes,
and $N_D$ the average number of descriptors per image. In the
original, each KD-tree contains $N_T N_D$ descriptors and each of the
$N_D$ query descriptors requires an approximate search in each of the
$N_C$ KD-tree structures. The accuracy of the approximate search is
controlled by the number of distance checks, $c$. The time complexity
for processing one query image under the original algorithm is $O(c
N_D N_C \log(N_T N_D))$. In our method, there is a single search
structure containing $N_D N_C N_T$ descriptors in which we search for
$k$ nearest neighbors (using $c$ distance checks, where $c \gg
k$). The time complexity for processing one query image under our
method is $O(c N_D \log(N_C N_T N_D))$. The $N_C$ term has moved
inside of the $\log$ term.

\section{Experiments and results}
We show results on both the Caltech 101 and Caltech 256 Datasets
\cite{Feifei2004, Griffin2007}. Each image is resized so that its
longest side is 300 pixels, preserving aspect ratio. We train using 15
and 30 images, common reference points from previously published
results. SIFT descriptors \cite{Lowe2004} are extracted on a dense,
multi-scale grid, and we discard descriptors from regions with low
contrast. We have attempted to match as closely as possible the
extraction parameters used by Boiman \etal.
\cite{Boiman2008}.\footnote{Our code and the feature sets used in our
  experiments will be made available for ease of comparison.} We
measure performance by the average per-class classification accuracy
(the average of the diagonal of the confusion matrix) as suggested by
\cite{Griffin2007}.

Boiman \etal \cite{Boiman2008} also introduced an optional parameter,
$\alpha$, that controls the importance given to the location of a
descriptor when matching. For all experiments, we fix $\alpha = 1.6$,
based on coarse tuning on a small subset of Caltech 101.

As discussed, we use FLANN \cite{Muja2009} to store reference
descriptors extracted from the labelled images in efficient approximate
nearest neighbor search structures.

\subsection{Tuning Local NBNN}
\begin{figure}
\centering
\includegraphics[width=0.5\linewidth]{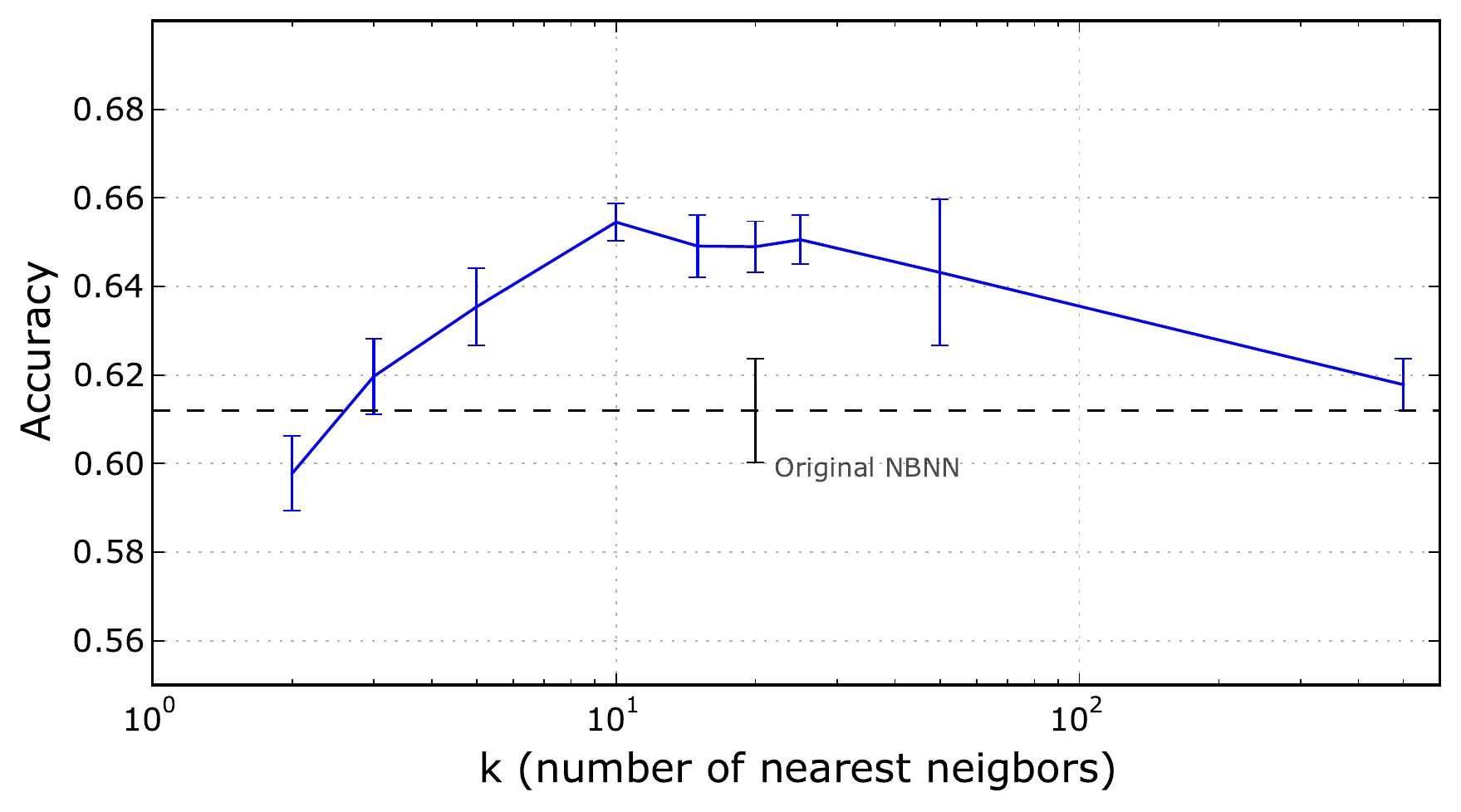}
\caption{The effect of changing $k$, the number of nearest neighbors
  retrieved from the merged index. Using only the local neighbors
  (about 10) results in optimal performance. The absolute performance
  numbers are lower than in our final results because we extracted
  fewer SIFT descriptors for this experiment.}
\label{fig:k_vs_accuracy}
\end{figure}

Figure \ref{fig:k_vs_accuracy} shows the effect of varying the
cut-off, $k$, that defines the local neighborhood of a descriptor. This
experiment shows that using a relatively low value for $k$ improves
performance. Using too low a value for $k$ hurts performance, and
using a much higher value for $k$ reverts to the performance of the
original NBNN.

We also demonstrate that this improved accuracy comes at a significant
time savings over the original. Instead of building 101 indices, local
NBNN uses a single index comprising all the training data, storing a
small amount of extra accessory data: the object class of each
descriptor.

We vary the computation afforded to both NBNN and local NBNN, and
track the associated classification accuracy. For local NBNN, we do a
search for 10 nearest neighbors, which returns an example from
approximately 7 of the classes on average. The selection of an
appropriately low number of nearest neighbors is important (see Figure
\ref{fig:k_vs_accuracy}).

To control the computation for each method, we control a parameter of
FLANN's approximate nearest neighbor search: the number of leaf-nodes
checked in the KD-trees. This also determines the accuracy of the
approximation. The higher the number of checks, the more expensive the
nearest neighbor searches, and the more accurate the nearest neighbors
retrieved. While FLANN does allow for auto-tuning the parameters to
achieve a particular accuracy setting, we fix the number of randomized
KD-trees used by FLANN to 4 so that we can control the computation
more directly. This setting achieves good performance with minimal
memory use.

\begin{figure}
\centering
\includegraphics[width=0.5\linewidth]{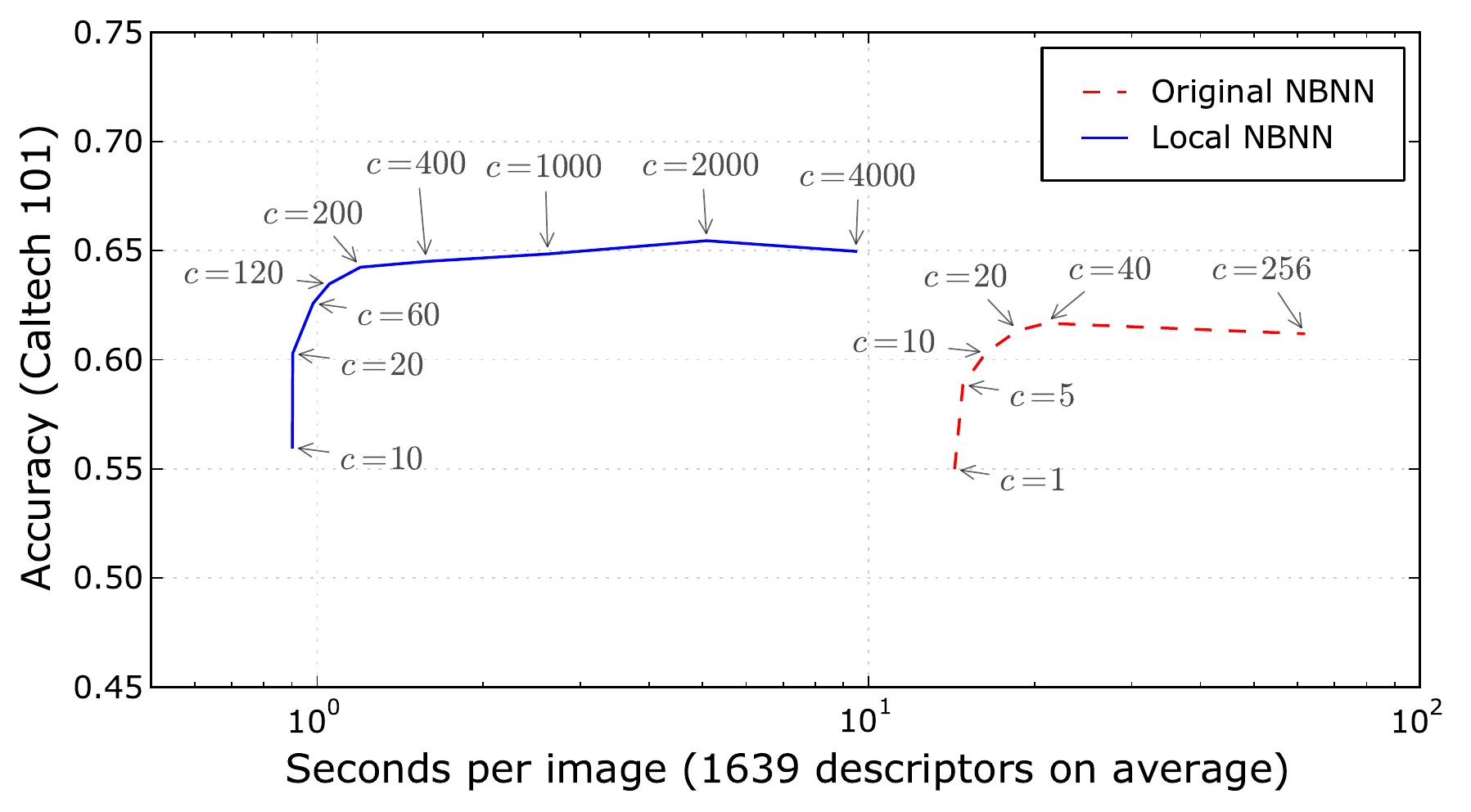}
\caption{Comparison of accuracy against computation for NBNN vs local
  NBNN. Computation is determined primarily by the number of distance
  checks ($c$ in this figure) allowed in the approximate nearest
  neighbor search. For 101 classes, even a single check in each of the
  101 indices is more expensive than one search with thousands of
  checks in the merged index due to the overhead of traversing each
  tree. These results were obtained on Caltech 101, using a sparser
  sampling of descriptors than in our final results.}
\label{fig:accuracy_vs_computation}
\end{figure}

Figure \ref{fig:accuracy_vs_computation} shows the results of this
experiment. There are significant improvements in both classification
accuracy and computation time. Looking in each of the 101 separate
class indices for just a single nearest neighbor in each, and checking
only one leaf node in each of those search structures was still slower
than localized search in the merged dataset. Even doing 1000, 2000, or
4000 leaf node checks in the merged dataset is still faster.

\subsection{Scaling to many classes}
\begin{figure}
\centering
\includegraphics[width=0.5\linewidth]{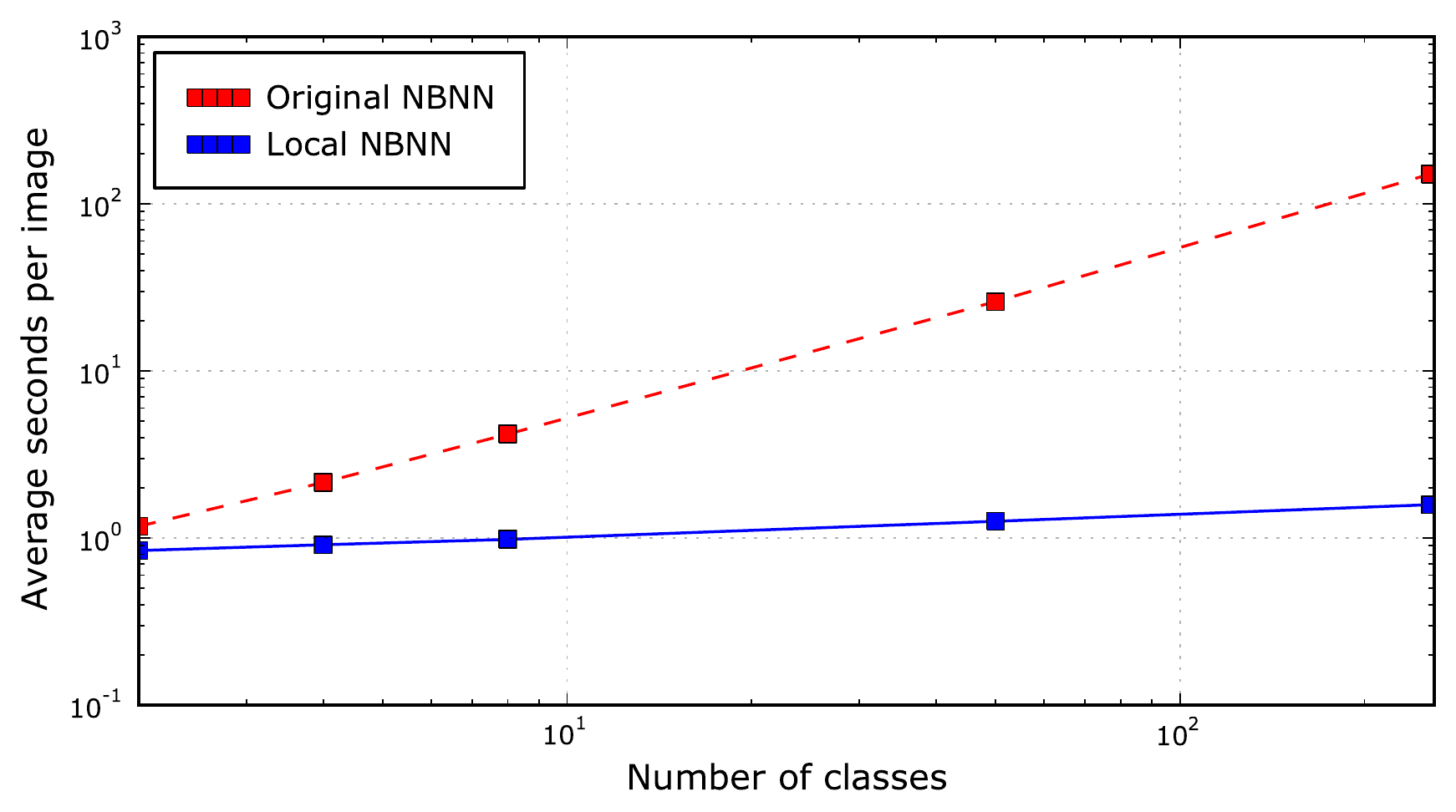}
\caption{We varied the number of categories from 2 up to 256 and plot
  the run time of the two methods. When classifying 256 categories,
  our method is 100 times faster than the original.}
\label{fig:scaling}
\end{figure}

Figure \ref{fig:scaling} further shows how the computation for these
two methods grows as a function of the number of classes in the
dataset. As new classes are added in our method, the depth of the
randomized KD-tree search structures increases at a log rate. As we
increase the number of classes to 256, local NBNN using the merged
dataset runs 100 times faster than the original. In the original
method, an additional search structure is required for each class,
causing its linear growth rate. This requires a best-bin-first
traversal of the each KD-tree. However, in the case where we query a
single search structure for 10-30 nearest neighbors, the
best-bin-first traversal from root to leaf happens only once, with the
remainder of the nearest neighbors and distance checks being performed
by backtracking. The preprocessing time to build the indices is almost
identical between the two methods.

\subsection{Comparisons with other methods}
Until now, no comparison has been done between NBNN and spatial
pyramid methods using the same base feature set. We show those results
in Table \ref{tab:comparison}. (Runtime for the original NBNN on
Caltech 256 was prohibitive, so we do not report those results.)

We choose to compare against two spatial pyramid methods. First, the
original model introduced by Lazebnik \etal
\cite{Lazebnik2006}. Second, a recent variant by Liu \etal
\cite{Liu2011} that takes advantage of local soft assignment in a
manner similar to our local cut-off, and that uses max pooling
\cite{Boureau2010a} rather than average pooling within each spatial
histogram bin. We trained a codebook of size 1024 for each of the
training set regimes (Caltech 101 with 15 and 30 training images,
Caltech 256 with 15 and 30 training images). Our spatial pyramid was 3
levels (1x1, 2x2, and 4x4 histogram arrangements). For classification,
we trained one-vs-all SVMs using the histogram intersection kernel
\cite{Lazebnik2006} and used a fixed regularization term for all
training regimes.

We also compare against some previously published figures for
NBNN. Notably, local NBNN gives the best performance of any NBNN
method to date.

While local NBNN (and NBNN) performs better the original spatial pyramid
model, it does not perform better than the model of Liu \etal. The
soft assignment avoids some of the information loss through
quantization, and the discriminative training step provides an
additional benefit.

The recent kernel NBNN of Tuytelaars \etal is a complimentary
contribution, and we suspect that the combinations of local NBNN with
the kernel NBNN would lead to even better performance. We hypothesize
that this combination would lead to NBNN matching or improving upon
the performance of state-of-the-art spatial pyramid methods.

There are other results using a single feature type that have higher
published accuracy on these benchmarks. For example, Boureau \etal
\cite{Boureau2011} show $77.1\%$ accuracy on Caltech 101 and $41.7\%$
on Caltech 256 with 30 training images, but they use a macro-feature
built on top of SIFT as their base feature, so that is not directly
comparable with our feature set. Combining different feature types
together would also yield higher performance as shown frequently in
literature \cite{Boiman2008, Tuytelaars2011}.

\begin{table*}\footnotesize
\begin{center}
\begin{threeparttable}
\begin{tabular}{ | l | c | c | c | c | }
\hline
& \parbox[b][0.6cm]{2.4cm}{\centering Caltech 101 \\ (15 training images)}
& \parbox[b][0.6cm]{2.4cm}{\centering Caltech 101 \\ (30 training images)}
& \parbox[b][0.6cm]{2.4cm}{\centering Caltech 256 \\ (15 training images)}
& \parbox[b][0.6cm]{2.4cm}{\centering Caltech 256 \\ (30 training images)} \\
\hline
\hline
\bf{Results from literature} & & & & \\
NBNN \cite{Boiman2008} & 65$\pm$1.14 & 70.4 & 30.5\tnote{1} & 37 \\
NBNN \cite{Tuytelaars2011} & 62.7$\pm$0.5 & 65.5$\pm$1.0 & - & - \\
NBNN kernel \cite{Tuytelaars2011} & 61.3$\pm$0.2 & 69.6$\pm$0.9 & - & - \\
\hline
\hline
\bf{Results using our feature set} & & & & \\
SPM (Hard-assignment, avg.-pooling)\tnote{2} & 62.5$\pm$0.9 & 66.3$\pm$2.6 & 27.3$\pm$0.5 & 33.1$\pm$0.5\\
SPM (Local soft-assignment, max-pooling)\tnote{3} & 68.6$\pm$0.7 & 76.0$\pm$0.9 & 33.2$\pm$0.8 & 39.5$\pm$0.4 \\
NBNN (Our implementation) & 63.2$\pm$0.9\tnote{4} & 70.3$\pm$0.6 & - & - \\
Local NBNN & 66.1$\pm$1.1 & 71.9$\pm$0.6 & 33.5$\pm$0.9 & 40.1 \\
\hline
\end{tabular}
\caption{Our local NBNN has consistent improvement over the original
  NBNN, outperforming all previously published results for NBNN using
  a single descriptor. We confirm NBNN outperforms the original
  spatial pyramid method, but is only competitive with the latest
  state-of-the-art variant.}
\begin{tablenotes}
\item [1] {Boiman \etal did not do an experiment with 15 images on this
  dataset. The 30.5 is an interpolation from their plot.}
\item [2] The original spatial pyramid match by Lazebnik \etal \cite{Lazebnik2006} (re-implementation).
\item [3] A recent variant of the spatial pyramid match from Liu \etal \cite{Liu2011} (re-implementation).
\item [4] {Our experiment using NBNN achieves $63.2\pm0.9$ compared to
  $65.0\pm1.14$ from \cite{Boiman2008}. The original implementation is
  not available, and we have had discussions with the authors to
  resolve these differences in performance. We attribute the disparity
  to unresolved differences in parameters of our feature extraction.}
\end{tablenotes}
\label{tab:comparison}
\end{threeparttable}
\end{center}
\end{table*}

\section{Conclusion}
We have demonstrated that local NBNN is a superior alternative to the
original NBNN, giving improved classification performance and a
greater ability to scale to large numbers of object
classes. Classification performance is improved by making adjustments
only to the classes found in the local neighborhood comprising $k$
nearest neighbors. Additionally, it is much faster to search through a
merged index for only the closest few neighbors rather than search for
a descriptor's nearest neighbor from each of the object classes.

Our comparison against spatial pyramid methods confirms previous
results \cite{Boiman2008} claiming that NBNN outperforms the early
spatial pyramid models. Further, while NBNN is competitive with the
recent state-of-the-art variants of the spatial pyramid, additional
discriminative training (as in the NBNN kernel of Tuytelaars \etal
\cite{Tuytelaars2011}) may be necessary in order to obtain similar
performance.

As new recognition applications such as web search attempt to classify
ever larger numbers of visual classes, we can expect the importance of
scalability with respect to the number of classes to continue to grow
in importance. For example, ImageNet \cite{Deng2009} is working to
obtain labelled training data for each visual concept in the English
language. With very large numbers of visual categories, it becomes
even more apparent that feature indexing should be used to identify
only those categories that contain the most similar features rather
than separately considering the presence of a feature in every known
category.

{\small
\bibliographystyle{ieee}
\bibliography{library.bib}
}

\end{document}